\documentclass{article} % For LaTeX2e
\usepackage{iclr2023_conference,times}

\usepackage{amsmath,amsfonts,bm}

\def\eqref#1{equation~\ref{#1}}
\def\1{\bm{1}}

\DeclareMathAlphabet{\mathsfit}{\encodingdefault}{\sfdefault}{m}{sl}
\SetMathAlphabet{\mathsfit}{bold}{\encodingdefault}{\sfdefault}{bx}{n}

\newcommand{\R}{\mathbb{R}}

\usepackage{hyperref}
\usepackage{url}

\usepackage[vlined,ruled]{algorithm2e}

\usepackage[utf8]{inputenc} % allow utf-8 input
\usepackage[T1]{fontenc}    % use 8-bit T1 fonts
\usepackage{hyperref}       % hyperlinks
\usepackage{url}            % simple URL typesetting
\usepackage{booktabs}       % professional-quality tables
\usepackage{amsfonts}       % blackboard math symbols
\usepackage{nicefrac}       % compact symbols for 1/2, etc.
\usepackage{microtype}      % microtypography
\usepackage{thm-restate}
\usepackage{microtype}
\usepackage{graphicx}
\usepackage{subfigure}
\usepackage{changes}

\usepackage{amsmath}
\usepackage{multirow}
\usepackage{hhline}
\usepackage{wrapfig}
\usepackage{floatrow}
\usepackage{caption}
\usepackage{enumitem}
\usepackage[export]{adjustbox}
\usepackage{amsmath,mathtools}

\usepackage{algorithmic}

\makeatletter
\newcommand{\raisemath}[1]{\mathpalette{\raisem@th{#1}}}
\newcommand{\raisem@th}[3]{\raisebox{#1}{$#2#3$}}
\makeatother

\newcommand{\uglad}{{\texttt{uGLAD}~}}
\newcommand{\ugladns}{{\texttt{uGLAD}}}
\newcommand{\glad}{{\texttt{GLAD}~}}
\newcommand{\tglad}{{\texttt{tGLAD}~}}
\newcommand{\tgladns}{{\texttt{tGLAD}}} % uGLAD-time

\title{Are \texttt{uGLAD}? Time will tell!}

\author{%\hspace{-5mm}
  Shima Imani~~~~~~~Harsh Shrivastava
  \hspace{0mm}\\
  \hspace{-3.55mm}
  \begin{tabular}{c}
      $\prescript{}{}{~\text{Microsoft Research, Redmond, USA}}$
  \end{tabular}
}

\newcommand{\Rho}{\mathrm{P}}

\iclrfinalcopy % Uncomment for camera-ready version, but NOT for submission.
\begin{document}

\maketitle

\begin{abstract}
We frequently encounter multiple series that are temporally correlated in our surroundings, such as EEG data to examine alterations in brain activity or sensors attached to a body to monitor movements. 
% Segmentation algorithms are designed to identify crucial points in a specified time series. Majority of these methods were primarily developed to analyse a univariate time series, the results for the more challenging problem of segmenting multivariate time series data still largely remains unsatisfactory.
Segmentation of multivariate time series data is a valuable technique for identifying meaningful patterns or changes in the time series that can signal a shift in the system's behavior. However, most segmentation algorithms have been designed primarily for univariate time series, and their performance on multivariate data remains largely unsatisfactory, making this a challenging problem.
% In this work, we devised a novel approach to do multivariate time series segmentation by making use of conditional independence (CI) graphs. 
In this work, we introduce a novel approach for multivariate time series segmentation using conditional independence (CI) graphs. CI graphs are probabilistic graphical models that represents the partial correlations between the nodes.
%CI graphs are a type of probabilistic graphical models that are based on underlying multivariate Gaussian assumption. 
We propose a domain agnostic multivariate segmentation framework `\tgladns' which draws a parallel between the CI graph nodes and the variables of the time series. Consider applying a graph recovery model \uglad to a short interval of the time series, it will result in a CI graph that shows partial correlations among the variables. We extend this idea to the entire time series by utilizing a sliding window to create a batch of time intervals and then run a single \uglad model in multitask learning mode to recover all the CI graphs simultaneously. As a result, we obtain a corresponding temporal CI graphs representation of the multivariate time series. We then designed a first-order and second-order based trajectory tracking algorithm to study the evolution of these graphs across distinct intervals. Finally, an `Allocation' algorithm is designed to determine a suitable segmentation of the temporal graph sequence which corresponds to the original multivariate time series. \tglad provides a competitive time complexity of $O(N)$ for settings where number of variables $D<<N$. We demonstrate successful empirical results on a Physical Activity Monitoring data.\footnote{\textit{Software}: \url{https://github.com/Harshs27/tGLAD}, {\href{https://sites.google.com/view/tglad/}{ website}}\\ \qquad\text{~~~ ~}*Both authors contributed equally} \\

\end{abstract}

\section{Introduction}

Time series segmentation is the process of dividing a time series into multiple segments, or sub-series, based on certain characteristics or patterns. 
% In general, time series segmentation can be used to identify and analyze patterns and trends in any data that is collected over time. 
Segmentation has many benefits, such as reducing a long time series into manageable sections to facilitate labeling by a human or machine annotator, and uncovering unexpected actionable patterns in data through exploration. For example, it can be used in finance to identify trends and patterns in stock prices, in marketing to analyze consumer behavior, and in healthcare to monitor patient vital signs. This helps to understand the underlying dynamics of the data and even make predictions about future events~\cite{omranian2015segmentation,yeh2016matrix,aminikhanghahi2017survey}.

There are numerous algorithms available for segmenting time series with majority of them primarily designed to handle the univariate case. If $N$ is the length of the time series, most algorithms have an expected time complexity of $O(N^2)$, however, some more recent algorithms have achieved an $O(N\log{N})$ time complexity with certain limiting approximations. Some time series segmentation methods are designed for specific domains, limiting their broader application. Additionally, some methods make assumptions about the semantic segments being well-defined, but they may not always align with real-world data and thereby hinder their effectiveness. The extensions suggested for these methods to handle multivariate data are non-trivial and often do not perform well in practice. Related works section covers them in detail.

Consider a small slice of a multivariate time series consisting of $D$ variables, say from $T_1$ to $T_{10}$ which contains no crucial segmentation points. For this slice, we can expect the correlation between the $D$ variables to be roughly the same throughout $\text{corr}^D(T_1)\sim \text{corr}^D(T_{10})$. Now, 
let's assume that there is a segmentation point at $T_{11}$. For instance, if we are monitoring the sensor data of an athlete, we can consider that at time $T_{11}$, the athlete switched activity from jogging to sprinting. We now expect that the correlations among the variables will change at the segmentation points, $\text{corr}^D(T_{10})\not\sim \text{corr}^D(T_{11})$. Our proposed framework, called \tgladns, is designed to efficiently detect this change of correlations which indicate segmentation points.

To realize this intuition, we identified a novel cross-domain application of sparse graph recovery for time series analysis. Briefly, given input variables and their samples, sparse graph recovery methods output a graph whose edges capture the direct dependencies among the variables. In our work, we focus on recovering special type of graphs, called the conditional independence (CI) graphs~\cite{shrivastava2022methods}. The CI graphs capture the partial correlations between the variables, which can be either positive or negative. Among the many different algorithms to recover CI graphs, we choose a recently developed state-of-the-art deep model called \ugladns~\cite{shrivastava2022a,shrivastava2022uglad}. Its multitask learning ability enables a single instance of the model to run on batch input and recover multiple graphs simultaneously, a property that paves way for the high efficiency of \tgladns. Although, one can theoretically utilize any algorithm under the larger umbrella of the sparse graph recovery methods, the methods section will highlight the key reasons which justifies our choice of using the combination of CI graphs and \ugladns.

% Fig.~\ref{fig:tglad-flow} outlines
The process followed by the \tglad framework for doing multivariate time series segmentation is as follows. We divide the time series into sub-sequences or batches and then run a CI graph recovery model \uglad that gives a corresponding temporal graphs. The nodes of CI graphs are the variables of the multivariate time series and they edges capture the partial correlation strength between the variables. In essence, we have distilled down some relevant information of the time series in the temporal CI graphs. As per our intuition, the instances where the consecutive CI graphs differ a lot in their correlations, those points of the temporal graphs will correspond to the segmentation points in the time series. We use this insight to develop our multi-step framework \tgladns. Thus, we developed efficient algorithms to capture the dynamics or the evolution pattern of the temporal CI graphs which in turn help us identify the segmentation regions in the original time series.  
%The time complexity of each of the individual step shown in Fig.~\ref{fig:tglad-flow} is of the order $O(N)$, thus giving an overall time complexity of \tglad framework as $O(N)$. 

Listing the key contributions of our work. Please note that we use the terms, time series and sequences, interchangeably throughout.
\begin{enumerate}[leftmargin=*,nolistsep]
    \item A novel cross-domain approach for multivariate time series segmentation based on using sparse graph recovery algorithms. 
    \item Efficient method to give linear $O(N)$ time complexity in terms sequence length for cases where the number of variables follow $D^3<<N$
    \item Provide explainability and transparency by giving insights into reasons for the segmentation.
    \item A domain agnostic framework that can applied for time series from various domains. %also be used to identify common sub-sequences occurring across the time series.
\end{enumerate}

% \harsh{list the key contributions.
% 1. cross-domain approach, combining sparse graph recovery algorithms for time series problem.
% 2. Explainability and transparency in predictions.
% 3. Can also be used to identify common segments across the time series. 
% 4. Especially designed for multivariate time series.
% 5. First method to give $O(N)$ 
% }

\section{Related works}

Our framework is a combination of the literature from time series segmentation and sparse graph recovery. So, we discuss relevant research from both of them to provide background knowledge.

\textbf{Segmentation Methods}. There are several time series segmentation methods available that use different approaches to segment a time series into different classes based on changes in its temporal shape patterns. We divide the existing methods into domain specific and domain agnostic ones.

\textit{Domain specific}. If one narrows down the scope to analysing time series to a specific field, specialized methods can be developed by utilizing the domain-specific insights. Survey in~\cite{lin2016movement} did a collective analysis of various such methods and one key insight they highlight is that for almost all the methods some background on the nature of the domain and motion is needed. Although, the recent observed trend is to develop domain agnostic approaches and we can find interesting techniques in this category. For example,  Automobile trajectories were studied in~\cite{harguess2009semantic}, electroencephalography data was analysed in~\cite{kozey2011validation}, electrical power consumption analysis in~\cite{reinhardt2013predicting}, music sequence analysis in~\cite{serra2014unsupervised}, biological time series in~\cite{omranian2015segmentation}, human motion segmentation was investigated in~\cite{lan2015automated,aoki2016segmentation,aminikhanghahi2017survey} among others.

\textit{Domain agnostic}. In attempt to design domain agnostic techniques for wider adaptability,  FLOSS (Fast Low-cost Online Semantic Segmentation)~\cite{gharghabi2019domain} was developed. It is a popular method which produces an Arc Curve (AC) that annotates the original time series with information about the likelihood of a regime change at each point in the series. The AC is used to identify segments with similar temporal shape patterns that are likely to belong to the same class and occur within close temporal proximity to each other. Another method called ESPRESSO (Entropy and ShaPe awaRe timE-Series SegmentatiOn)~\cite{Deldari2020ESPRESSOEA}, is a hybrid approach that uses both shape pattern and statistical distribution of time series to segment time series data. ESPRESSO uses a modified version of FLOSS by incorporating the Weighted Chained Arc Curve to capture the density of pattern repetition with time. Recently proposed ClaSP (Classification Score Profile) is a self-supervised time series segmentation method that uses overlapping windows to split time series into hypothetical partitions. For each partition, a binary classifier is trained and evaluated using cross-validation. The degree of self-similarity is recorded for each offset and then the classification score profile is computed, which is ultimately used for segmenting time series data~\cite{Schfer2021ClaSPT,ermshaus2022clasp}. Other relevant methods include~\cite{imanitime2cluster, Castellini2020TimeSS, Lu2020SegmentationOM, Machn2017SimilarityBasedSO}.

\begin{figure}[h]
    \centering 
    \includegraphics[width=135mm]{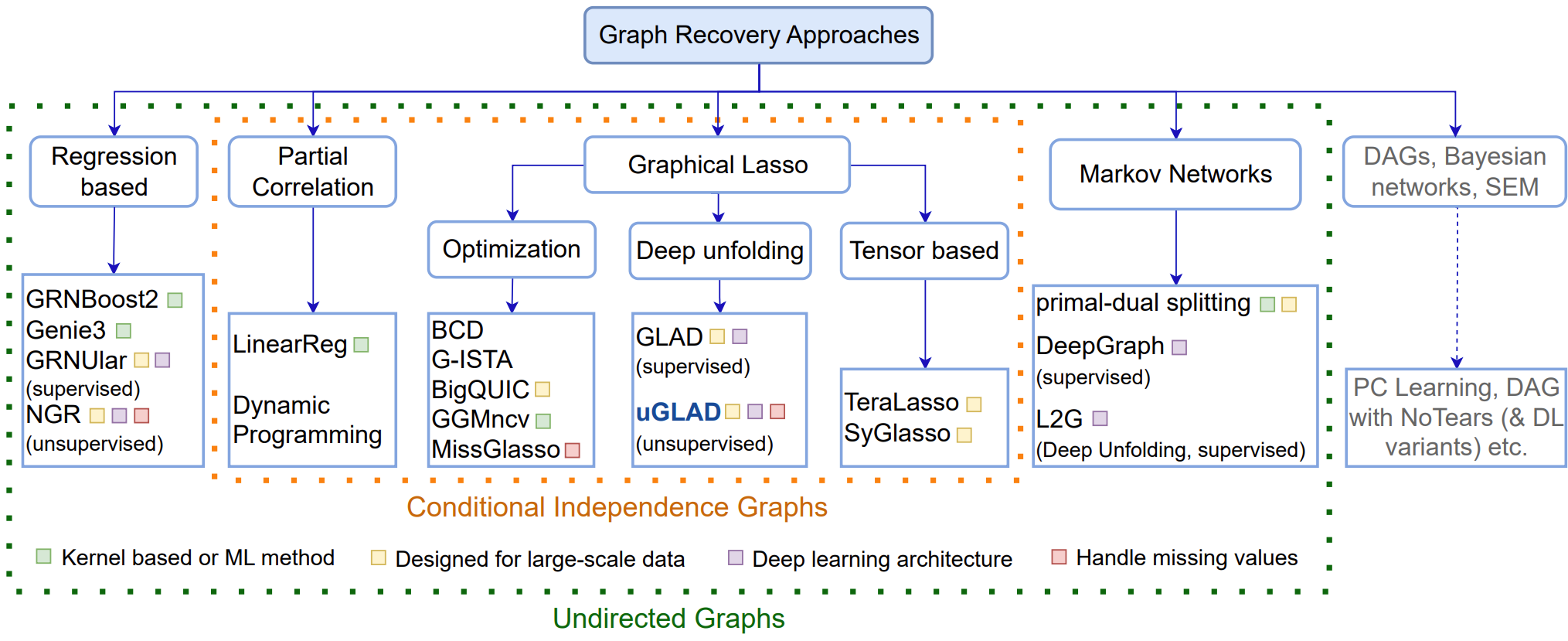}
    \caption{ \textit{Overview of Sparse Graph Recovery methods}. We focus on methods that recover undirected graphs which capture direct dependence among their nodes or features. \tglad framework utilizes a recently developed deep model, \ugladns, that outputs a conditional independence graph between in the features. Our framework can potentially use other methods and will be interesting topic for future explorations. (partly borrowed from~\cite{shrivastava2023neural})}
    \label{fig:sparse-methods}
    % \vspace{-5mm}
\end{figure}

\textbf{Sparse graph recovery}. Given data with $D$ features and $M$ samples as input, the aim of the sparse graph recovery methods is to obtain a probabilistic graphical model~\cite{koller2009probabilistic} that potentially shows sparse connections between the $D$ features. We focus on methods that recover undirected graphical models, refer Fig.~\ref{fig:sparse-methods}. Sparse graph recovery methods have been used for various applications like gene regulatory network discovery~\cite{van2010inferring,haury2012tigress,moerman2019grnboost2,shrivastava2020grnular,aluru2021engrain,shrivastava2022grnular}, understanding Digester functioning to increase Methane yield~\cite{shrivastava2022uglad}, extracting insights from an Infant mortality data~\cite{shrivastava2022neural,shrivastava2023neural}, studying autism by analysing brain sensory signals~\cite{pu2021learning} among many others.

\textit{Conditional Independence graphs}. The edges of a CI graph show the partial correlation between the nodes or features. The partial correlation can be considered as capturing direct dependency between the features as it is the conditional probability of the features under consideration given all the other features. Refer inner block in orange of Fig.~\ref{fig:sparse-methods}. Popular formulations of recovering CI graph include optimizing the graphical lasso objective~\cite{friedman2008sparse,rolfs2012iterative} which include deep models like~\cite{shrivastava2019glad,shrivastava2020using,shrivastava2022a,shrivastava2022uglad} or dynamic programming based approach to directly evaluate the expression of partial correlations. Survey~\cite{shrivastava2022methods} formalizes the definition of CI graphs, categorizes various methods that recover such graphs, describe and compares their performance, provide their implementation details and discuss their applications. It is a good entry point to understanding the umbrella of methods that recover CI graphs. The method by~\cite{hallac2017network} they utilized temporal graphs to understand dynamics of systems which is similar idea as ours but was not developed for time series segmentation settings. 
% \cite{hallac2019greedy} introduced a method called Segmented Gaussian Model. They define a top-down approach that runs for a fixed number of breakpoints. Their base algorithm scales as square of the sequence length and provide a heuristic algorithm that approximates the optimization to run in linear time.

% https://deepai.org/publication/the-opportunity-challenge-a-benchmark-database-for-on-body-sensor-based-activity-recognition

% http://archive.ics.uci.edu/ml/datasets/OPPORTUNITY+Activity+Recognition

\section{Methods}
We introduce the necessary definitions and notations to facilitate our discussions followed by the steps followed by the \tglad framework. 

\subsection{Definitions}

A \textbf{multivariate time series} \(T\) of length \(N\) and dimension \(D\) is a sequence of real-valued vectors 
\[T = {t_1, t_2, \ldots, t_N}, \text{ where } t_i \in \mathbb{R}^D\]

A \textbf{Subsequence} is defined as a local section of a time series that consists of a continuous subset of its values. A subsequence $T_{i,M}$ of a time series $T$ is a continuous subset of the values from $T$ of length $M$ starting from position $i$. Formally, $T_{i,M} = t_i, t_{i+1},…, t_{i+M-1}$, where $1$ $\le$ $i$ $\le$ $N-M+1$.

In order to extract continuous subsequences from time series, we utizlize the \textbf{stride length} shifting to determine the next subsequence. In time series data, the stride length is the number of data points by which we shift the starting position of the current subsequence to extract the position of the next subsequence. For example, a stride length $s$ means that if the current subsequence is located at \(T_{i,M}\) where \(i\) where is the starting position of the subsequence from \(T\) with length \(M\), then the next subsequence is \(T_{i+s,M}\) with the starting position at \(i+s\).

\subsection{\tglad framework}

\begin{figure}[h]
\centering 
\includegraphics[width=135mm]{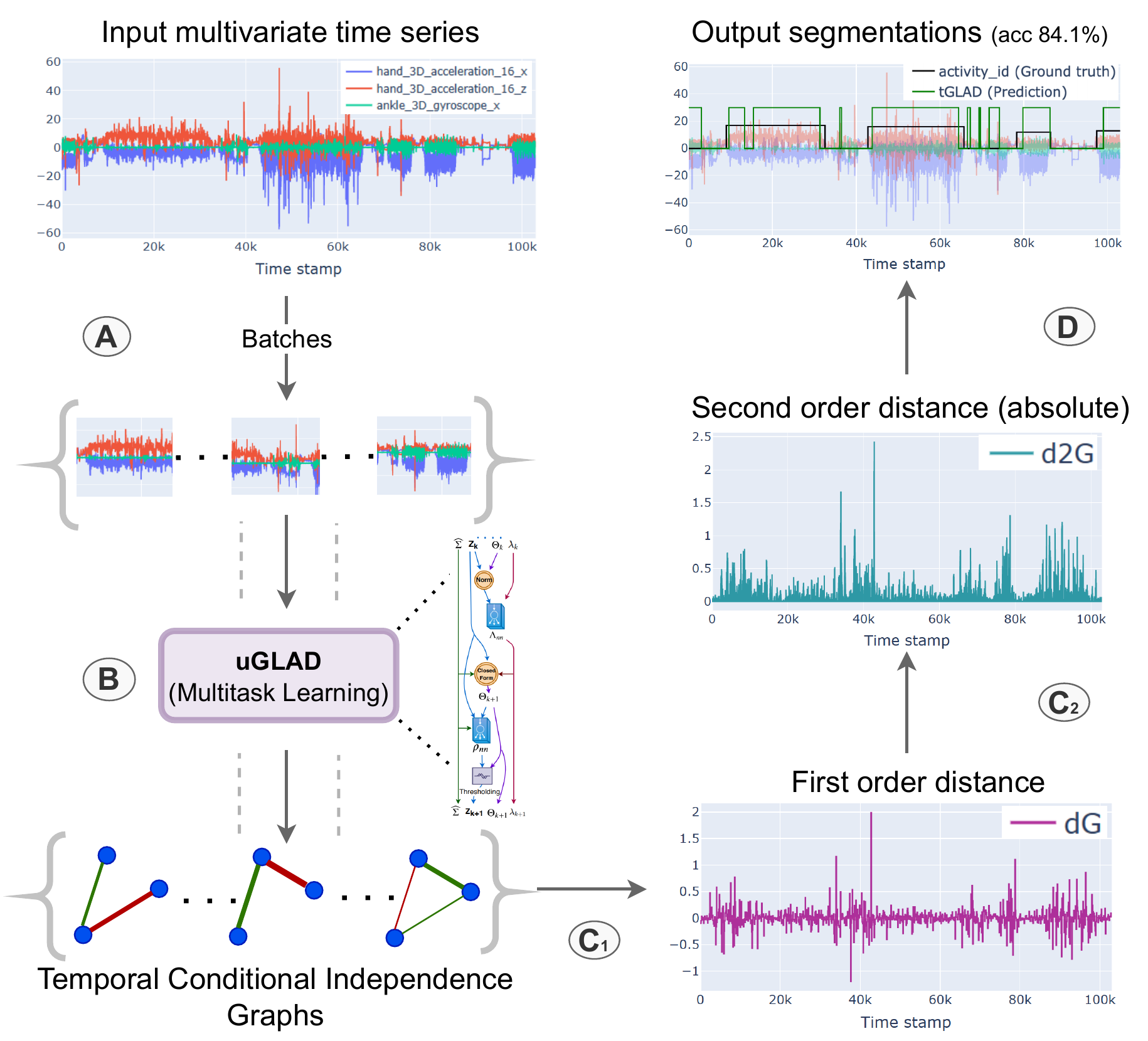}
\caption{\textit{\tglad framework}. (\textbf{A}) The time series is divided into multiple intervals by using a sliding window to create a batch of intervals. (\textbf{B}) Run a single \uglad model in multitask learning (or batch) mode setting to recover a CI graph for every input batch. This gives a corresponding set of temporal CI graphs. The entire input is processed in a single step as opposed to obtaining a CI graph for each interval individually. (\textbf{C$_1$}) Get the first order distance, $dG$ sequence, of the temporal CI graphs which captures the distance between the consecutive graphs. This is supposed to give higher values at the segmentation points. (\textbf{C$_2$}) Again take a first order distance of the sequence in the previous step and then its absolute value to get $d2G$ sequence, which further accentuates the values at the segmentation points. (\textbf{D}) Apply a threshold to zero out the smaller values of $d2G$ and identify the segmentation blocks using an `Allocation' algorithm.}
\label{fig:tglad-flow}
% \vspace{-10mm}
\end{figure}

Figure~\ref{fig:tglad-flow} enumerates the steps followed by \tglad to do multivariate time-series segmentation. The details for each of these steps are given below.

\textbf{(A) Identifying variables and prepare batch input for sparse graph recovery}

For all the variables in the given multivariate time series, basic preprocessing is done which includes missing value imputation using a forward filling algorithm. The data is now partitioned into small chunks using a fixed window size $M$ and stride length $s$ and runs over the entire time series. The window size determined based on the approach suggested in~\cite{imani2021multi}. We now end up with $B=(N-M+1)/s$ batches, with each having $M$ samples for $D$ variables. The input to the graph recovery algorithm will be the batch of samples, represented as a tensor of size $X\in\mathbb{R}^{B\times M\times D}$.

% \shima{
% To prepare the dataset for analysis using the \ugladns algorithm, missing values are filled using forward filling as a preprocessing step. Then, the data is partitioned into multiple time series using a fixed window size and stride length, with the window size determined based on the approach suggested in~\cite{?}. The resulting time series are grouped into batches as shown in Fig.~\ref{fig:tglad-flow}(A).
% }
% % \harsh{Based on the window size, stride length and batch size, create a batch of input matrices that describes chunks of the time series. These set of matrices act as the batch input data for \ugladns.} 

\textbf{(B) Obtaining the temporal Conditional Independence graphs}

\begin{wrapfigure}[15]{R}{0.47\textwidth}
% \begin{figure}
\vspace{-4mm}
    \centering
\begin{algorithm}[H]
  \DontPrintSemicolon
  \SetKwProg{Fn}{Function}{:}{}
\SetKwFor{uFor}{For}{do}{}
\SetKwFor{uIf}{If}{then}{}
  \SetKwRepeat{Do}{do}{while}
  \SetKwFor{ForPar}{For all}{do in parallel}{}
  \SetKwFunction{predict}{get-segments}
    \Fn{\predict{$\textbf{d2G}$, Z=5}}{
        $B \gets len(\textbf{d2G})$\;
        $labels \gets [1]\times B$\;
        % $\textbf{pred}\gets \textbf{1}$, \;
        % $\textbf{N}\gets \textbf{Length(pred)}$, \;
        \tcc{Removing noise}
        $\textbf{d2G} < 0.5 = 0$ \;
        \tcc{The window size is M}
        % $\textbf{win}\gets 5\textbf{w}$, \;
         \uFor{$i\gets0 ~\KwTo ~B$}{
            \uIf{d2G[i] > 0}{
            $start = max(0, i - M\cdot Z)$  \;
            $end = min(B - 1, i + M\cdot Z)$ \;
            $labels[start:end] \gets 0$\;
            }
        }
    \KwRet $labels$
    } 
    % \vspace{2mm}
\caption{Allocating segments}\label{algo:pred}
\end{algorithm}%\vspace{-5mm}
% \end{figure}
\end{wrapfigure}

The aim of the sparse graph recovery algorithm is to run on the input from step (A), denoted by $X$ and output corresponding set of graphs, whose adjacency matrix is represented here by the tensor $\Rho\in\mathbb{R}^{B\times D\times D}$. There are 2 key requirements from any such method, namely (1) The resultant graph should capture direct dependencies between the features (2) The method should be efficient. We chose a combination of CI graphs and \uglad model keeping in mind the desiderata desired. 

\textit{Why CI graphs}? CI graphs capture partial correlations between the features which model direct dependencies between them. The nodes are the features and the edge weights carry the partial correlation value that lies in the range $[-1,1]$. This additionally provides us with the positive or negative correlation information, which helps us later in determining the relevant features that result in a segmentation prediction as well as provide explainability and transparency to our framework.

\textit{Why \ugladns}? Introduced in~\cite{shrivastava2022a}, \uglad is a deep-unfolding (or unrolled algorithm) based model which is an unsupervised extension of the \glad~\cite{shrivastava2019glad} model. These models are based on the optimization of the graphical lasso objective which assumes that the observed data comes from an underlying multivariate Gaussian distribution. Owing to the deep-unfolding done based on the Aternating Minimization updates and then expressiveness provided by the neural network based parameterization, these models are shown to better capture the tail-distribution points and also improve sample complexity results. Apart from the theoretical advantages and performance improvements over the other CI graph recovery methods, \uglad is efficient as well. The tensor based implementation of \uglad allows it to do multitask learning. This enables a single model to recover the entire batch of data simultaneously. We want to point out that we consider the sample data within a window size follow i.i.d. setting for the multivariate Gaussian assumption to work.

We run \uglad in `batch mode' to obtain all the underlying precision matrices at once, $\theta\leftarrow$\ugladns($X$), where $\theta\in\R^{B\times M\times D}$. The calculation of the partial correlation matrix $\Rho$ is straightforward from $\Theta$, refer~\cite{shrivastava2022methods}. The parameter sharing across these different tasks helps maintain robustness against noisy data and facilitates transfer learning. We thus obtain a series of temporal CI graphs, represented by the adjacency matrices $\textbf{G}=[G_1, G_2, \cdots, G_B]\in\mathbb{R}^{B\times D\times D}$ using \textbf{$\Rho$}. Each entry of the adjacency matrix is equal to the partial correlation value, $G_b[p,q]=\rho(D_p, D_q)$ for the $b^{th}$ batch and $D_k$ represent the $k^{th}$ time series variable. The temporal graphs can be seen as distilling some relevant information from the original multivariate time series data in form of graphs.

\textbf{(C) Towards segmentation of the corresponding temporal CI graphs}

Our formulation is based on the assumption that the key signals needed to successfully segment the original time series are captured in the corresponding temporal graphs and that the correlation among the features are informative enough for the task. So, if we are able to segment the temporal graphs, we can map the segmentation to the original time series. 

($C_1$) We compute the first-order distance sequence $dG\in\mathbb{R}^B$ by finding the distance of the consecutive graphs in the temporal graph series $\textbf{G}$. For each entry $b\in B$ of $dG$, we measure the distance between its recovered graph and the next neighbor as
% \[dG[i] = \operatorname{distance}(G_i, G_{i+1}) = \sum_{e}(weight_{i,e} - weight_{i+1, e}) ~~\forall{e\in{G_i\text{.edges}\cup G_{i+1}\text{.edges}}} \] %\texttt{ where e} \in \texttt{enumerate-all-edges}
\[dG[b] = \operatorname{distance}(G_b, G_{b+1}) = \sum_{p,q}(G_b[p,q] - G_{b+1}[p,q]) ~~\forall{p,q}\in\{1,\cdots, D\} \] 

where weights are the partial correlation values of the edges of the CI graphs under consideration. 

($C_2$) Given the sequence $dG$, next we compute the second-order distance sequence $d2G$ by applying the following distance operation 
\[d2G[b] = \operatorname{abs}\left(dG[b] - dG[b-1]\right),~~ \forall{b}\in(1,B)\]
The first-order distance measures the change between each recovered graph and its next neighbor, while the second-order distance highlights potential segmentation points. While there are other distance metrics that can potentially be used, in our experiments, we found that the first-order and second-order distances described above worked well for detecting segmentation points. The output of this trajectory tracking step is the $d2G$ sequence.

% \harsh{Explain the derivative of temporal CI graphs. For this we need to define a distance metric between 2 CI graphs. Briefly state the different choices of functions that can be potentially chosen. We experimentally found that even the different between the corresponding adjacency matrices works well. The output of this step is the dG array}

\textbf{D. Allocation algorithm for obtaining the final segmentation}

We develop an `Allocation' algorithm to obtain the final segmentation points from the $d2G$ sequence. We first filter out small noises in $d2G$ by applying a conservative threshold. We then traverse the sequence $d2G$ sequentially and mark the start of a segmentation a new block if we observe a non-zero value. We also disregard any changes in behavior or segmentation points that occur in less than $Z$ times the window size (M), otherwise the segmentation size will be significantly smaller than the window size and we will not be able to catch it. We usually choose $Z\sim 5$ in our experiments. The allocation process (Alg.~\ref{algo:pred}) reads the $d2G$ sequence and predict the \tglad segmentation scores.

% \harsh{We now want to design a trajectory tracking algorithm for the temporal graphs by leveraging the dG array obtained. We now create a d2G array, which takes the derivative (or difference) of the dG array. This d2G array now outlines the segmentation blocks in our original multivariate time series. }

\subsection{Time complexity analysis of \tglad}

We analyse the time complexity of each of the steps followed by the \tglad framework below.
\begin{itemize}[leftmargin=*]%,nolistsep]
    \item \textit{(A)} Creation of batches will require a single full scan of the time series, so complexity is $O(N)$.
    \item \textit{(B)} the time complexity of this step will consist of the input covariance matrix creation $O(N\cdot D^2)$, and then running \uglad in batch mode. For a single input, \uglad runs in $O(D^3\cdot E)$, so for $B$ batches the sequential runtime will be $O(B\cdot D^3\cdot E)$. Since, we can process batches in parallel with \uglad batch mode, in practice we observe significantly less runtime. The worst case scenario will be when $B\rightarrow N$, giving time complexity as $O(N\cdot D^3\cdot E)$.
    % Input covariance calculation  and \uglad runs in $O(B\cdot D^3\cdot E)$
    % Running \uglad has a time complexity of $O(N/B\cdot D^3\cdot E)$, where $E$ is the number of epochs for convergence, typically in the range of (1K, 5K). A single \uglad model processes $B$ batches at a time and is optimized for running on parallel computes and GPUs. We often want to keep $B$ as large as possible, so that each of these individual tasks benefits from the parameter sharing. We keep the stride length $s=1$ as a conservative choice. If memory (RAM) permits, we tend to choose $B\sim N$, so the time complexity becomes $O(D^3\cdot E)$. If $D$ becomes large, processing the entire data in a single batch will become increasingly difficult, so the $N/B$ ratio will increase. The worst case time complexity estimate will be $O(N\cdot D^3\cdot E)$ the for the extreme case, where we process the graphs sequentially, i.e. $B=1$.
    \item \textit{($C_1$)} The first order distance function goes through the entire length of the temporal graph sequence and each time enumerate all possible edges between graphs having $D$ nodes. So, it has a time complexity of $O(N\cdot D^2)\sim O(N)$. 
    \item \textit{($C_2$)} Creation of $d2G$ will require a single full scan of the $dG$, so complexity is $O(N)$.
    \item \textit{(Allocation algorithm)} Scans the $d2G$ array once, so complexity is $O(N)$.
\end{itemize}

The overall time complexity of the \tglad framework in cases where the number of variables are not high, $D^3<<N$, is $O(N)+O(N\cdot D^3\cdot E)+O(N\cdot D^2) \sim O(N)$. The worst case time complexity, where the number of variables are so large 
% which leads to $B\sim N$ number of batches close to the length 
that we cannot leverage the power of the multitask learning in batch mode of \ugladns, is $O(N)+O(N\cdot D^3\cdot E)+O(N\cdot D^2) \sim O(N\cdot D^3)$.

\section{Experiments}
We evaluate the \tglad framework on a real world body sensor dataset. Since, it is a novel framework, we conduct several design choices experiments to understand their impact on \tgladns's performance. 

\subsection{PAMAP2 Dataset}
To get a realistic sense about the effectiveness of our approach, we conducted experiments on the PAMAP2 Physical Activity Monitoring dataset~\cite{reiss2012introducing}. This dataset captures sensor data from multiple participants engaging in a variety of physical activities, making it a valuable resource for activity recognition and algorithm development. Our analysis was primarily based on multi-dimensional time series with the following three signals: the hand acceleration signal in the x-axis and z-axis, and the ankle gyroscope signal in the x-axis, which allowed us to examine the movements and rotations of the hand and ankle during physical activity.

Fig.~\ref{fig:tglad-flow} (A \& D) shows a three-hour segment of this data collected from one of the participants, highlighting their physical activities such as ironing (44 minutes), vacuum cleaning (42 minutes), and stair activity (ascending 15 minutes, descending 10 minutes), as well as periods of inactivity (transient). Fig.~\ref{fig:tglad-flow} additionally shows all the steps of the \tglad framework followed in order to segment the data. % accuracy of 84.1\% 

% For example, the plot shows that the participant started with a period of inactivity lasting about 17 minutes, followed by ironing for approximately 44 minutes. After a brief rest of 19 minutes, the participant engaged in vacuum cleaning for about 42 minutes , followed by another period of rest lasting 23 minutes. Finally, the participant descended stairs for approximately 15 minutes, took another rest period of about 15 minutes, and then descended the stairs for another 10 minutes.

% \harsh{Explain the dataset. Variables and what they mean. Also show a plot where the purple plot shows the segmentation and the 3 signals of acceleration are present. We present our results for this data. Our aim we these experiments is exploring the abilities of various components of \tgladns.}

\subsection{Results}
% \begin{figure}%[b]
% \centering 
% \includegraphics[width=120mm]{figures/correlationmatrix.png}
% \caption{Correlation matrix corresponding to the  multivariate time series of Figure \ref{fig:tglad-flow}.}
% \label{fig:correlation}
% \end{figure}

We chose accuracy as the metric to evaluate the segmentation performance. The accuracy is measured as the penalty for mislabeling the segmentation. For the ground truth time series, we put label=$1$ whenever an activity occurs and at every segmentation point where there is no activity, we switch the label=$0$. For the prediction labels, we consider the $d2G$ sequence obtained from Fig.~\ref{fig:tglad-flow}(D) and use the Allocation technique describe in Alg.~\ref{algo:pred}, with parameter $Z=5$.

\begin{figure}%[b]
\centering 
\includegraphics[width=100mm]{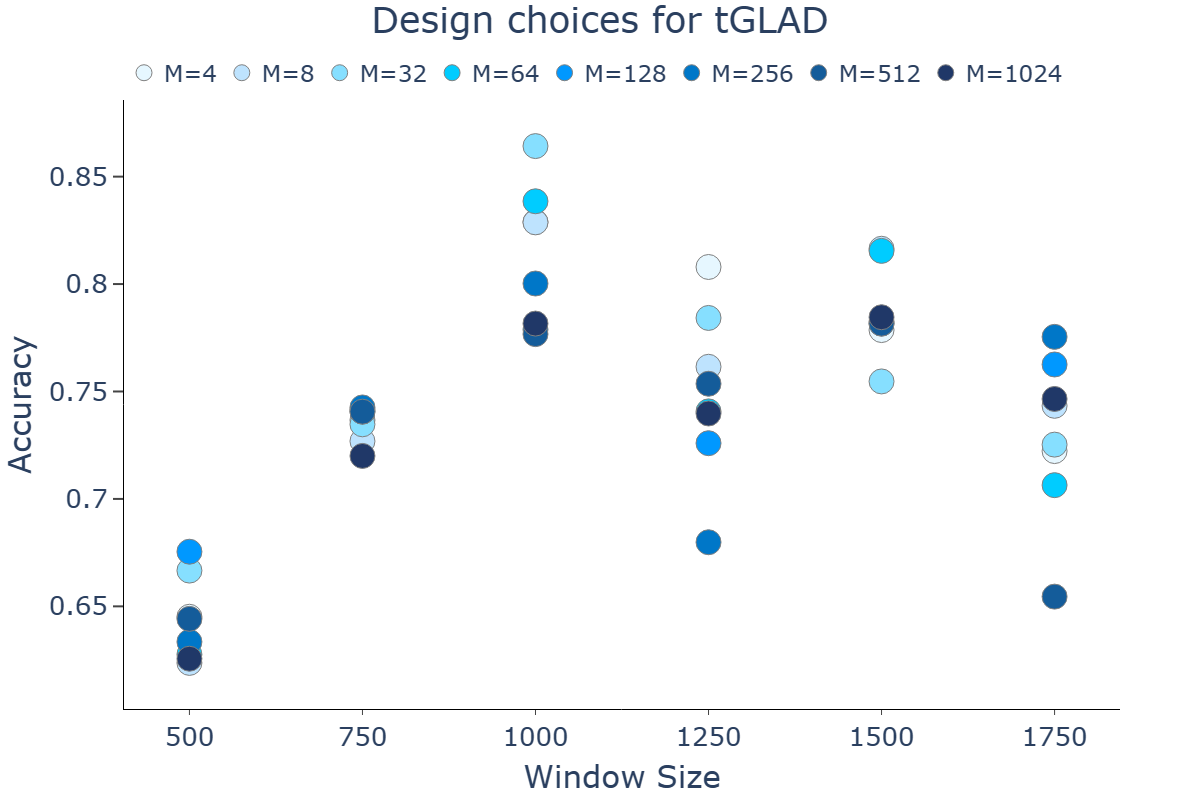}
\caption{\textit{Design choices for \tgladns}. Examining the segmentation accuracy on the PAMAP2 dataset which records body sensor data. We vary the window size on the x-axis and for each window size, we evaluate the performance for varying batch sizes (M). The stride length was fixed at 100 for all the experiments.}
\label{fig:window_batch_accuracy}
\end{figure}

We achieved an accuracy of 84.1\% for the PAMAP2 dataset using a window size of 1000 , batch size of 64 and stride length of 100, indicating that the \tglad framework is effective for physical activity monitoring based time series. The window size is the chunk of the time series considered at a time for processing CI graphs, so it is an important parameter to be chosen while running our framework. The batch size is the number of graphs that are recovered by a single \uglad model. As we are using multitasking, the parameters of the model are shared among the graphs within a batch, hence this is also an important parameter that can affect \tgladns's performance. Small batch size will lead to increased runtime as more number of batches to process, less robust to noise but more accurate graph recovery, while on the other hand, higher batch size will be efficient in term of runtime and robustness to anomalies, but since it has to recover graphs that are potentially sampled from different underlying distributions, the accuracy might take a hit. So, it is imperative that we do a study on effect of these design choices. Thus, in order to gain insights into the performance of \tglad with respect to the batch size and window size, we explored the impact on the segmentation accuracy by doing a grid plot over a range of size choices, as illustrated in Fig.~\ref{fig:window_batch_accuracy}.

Analysing the results indicate that changes in batch size and window sizes do not significantly affect the accuracy of the segmentation. If we consider any fixed window size, we do not see much variance in the performance over different batch sizes, that suggests a good graph recovery performance of the \uglad model. Thus, we can potentially increase the batch size for faster runtimes, without compromising much on the accuracy. Lots of research has been done on the choice of window size, with some methods being more sensitive than others. We do see variance in the performance of \tglad with change in window size, still the results suggest that a reasonable window size can be chosen to achieve a satisfactory segmentation label. The choice of the window size also depends considerably on the type of data as well.

% \harsh{Fig can show the 3 big green correlation matrices with varying window length.}

% \harsh{[optional] For a particular window length, if we change the stride length?}

% \harsh{[optional] Effect of batch size on the output?}

% \harsh{Varying window lengths, show figures of that orange-blue curves which outline the segmentation blocks. }

% expt 1
% window_size = 200
% batch_size = 20
% stride_length = 20

% expt 2
% window_size = 1000
% batch_size = 20
% stride_length = 100

\section{Conclusions}
We introduce a domain agnostic multivariate time series framework called \tgladns. It is a novel cross-domain approach that maps the original time series to a corresponding temporal graph representation which makes the problem of finding segmentation easier and efficient. The choice of a recently developed deep model \uglad for recovering conditional independence graphs gives the much needed efficiency to our framework. We identified a unique use of the multitask learning ability of \uglad model which also makes the case of batch learning in sparse graph recovery models more lucrative. Additionally, from the plethora of graph choices available, this work also narrowed down the type to conditional independence graphs. The CI graphs capture the intuition that correlation among the multivariate timeseries variables will change significantly at the segmentation points. We demonstrate successful segmentation results on the challenging PAMAP2 dataset, with achieving an accuracy of 84.10\% along with performing a parameter exploration study.  

\subsection{Future Work}

We have plans to pursue two directions of research for expanding \tgladns. Firstly, we will investigate the potential for segmenting univariate time series data using the \tglad framework. our approach consists of `smartly' converting the 1D sequence to multidimensional time series and then use the \tglad framework. This approach seems promising due to its high efficiency in terms of time complexity and hopefully good segmentation accuracy. Secondly, we aim to extend the \tglad framework to work in real-time or online settings. This will require adapting the framework and evaluating the trade-offs between computational efficiency and segmentation accuracy. The results of this research could have significant implications for fields such as finance, healthcare, and industrial monitoring.

Multivariate time-series segmentation, especially \tglad algorithm can be potentially applicable to the video representation and video generation domains. Videos can be interpreted as a sequential collection of frames~\cite{oprea2020review}. Recently, graphical methods that model the feature interactions within a frame and fitting temporal processes over the sequence of frames provide a strong alternative approach for many different video tasks~\cite{shrivastava2024methods}. Examples include video segmentation \& object detection, action recognition, prediction of interaction activity~\cite{bodla2021hierarchical,jiao2021new,zhou2022survey,saini2022recognizing}. Multivariate time series segmentation help in identifying similar segments and draw contrast between different window time frames. The \tglad algorithm can be helpful for efficient video prediction task by continuous multi-dimensional processes~\cite{shrivastava2024video1}. It can also find potential applications for efficient video prior representation learning~\cite{shrivastava2024video2,shrivastava2024video3} and the challenging task of generating diverse video frames~\cite{denton2018stochastic,shrivastava2021diverse,shrivastava2021diversethesis}. Conditional Independence graphs have found their way into applications with text mining tools~\cite{roche2017valorcarn,fize2017geodict,antons2020application}, where \tglad can be leveraged to improved interpretability analysis over a batch of text documents.

% \harsh{
% \begin{enumerate}
%     \item Add 1D time series analysis and possible reduction of multivariate to 1D variable. 
%     \item Extension to online settings. 
% \end{enumerate}
% }

% \subsection{Notes, to be deleted}

% \harsh{
% \begin{enumerate}
%     \item Study the trajectories of features. Trajectory detection algorithm on how the graph evolves over the intervals. 
%     \item add the edge case, where the correlations remain the same and the segmentation occurs.
%      \item add runtime, compute used to run expts. 
% \end{enumerate}
% }

\bibliography{citations}

\begin{thebibliography}{53}
\providecommand{\natexlab}[1]{#1}
\providecommand{\url}[1]{\texttt{#1}}
\expandafter\ifx\csname urlstyle\endcsname\relax
  \providecommand{\doi}[1]{doi: #1}\else
  \providecommand{\doi}{doi: \begingroup \urlstyle{rm}\Url}\fi

\bibitem[Aluru et~al.(2021)Aluru, Shrivastava, Chockalingam, Shivakumar, and
  Aluru]{aluru2021engrain}
Maneesha Aluru, Harsh Shrivastava, Sriram~P Chockalingam, Shruti Shivakumar,
  and Srinivas Aluru.
\newblock Engrain: a supervised ensemble learning method for recovery of
  large-scale gene regulatory networks.
\newblock \emph{Bioinformatics}, 2021.

\bibitem[Aminikhanghahi \& Cook(2017)Aminikhanghahi and
  Cook]{aminikhanghahi2017survey}
Samaneh Aminikhanghahi and Diane~J Cook.
\newblock A survey of methods for time series change point detection.
\newblock \emph{Knowledge and information systems}, 51\penalty0 (2):\penalty0
  339--367, 2017.

\bibitem[Antons et~al.(2020)Antons, Gr{\"u}nwald, Cichy, and
  Salge]{antons2020application}
David Antons, Eduard Gr{\"u}nwald, Patrick Cichy, and Torsten~Oliver Salge.
\newblock The application of text mining methods in innovation research:
  current state, evolution patterns, and development priorities.
\newblock \emph{R\&D Management}, 50\penalty0 (3):\penalty0 329--351, 2020.

\bibitem[Aoki et~al.(2016)Aoki, Lin, Kuli{\'c}, and
  Venture]{aoki2016segmentation}
Takashi Aoki, Jonathan Feng-Shun Lin, Dana Kuli{\'c}, and Gentiane Venture.
\newblock Segmentation of human upper body movement using multiple imu sensors.
\newblock In \emph{2016 38th Annual International Conference of the IEEE
  Engineering in Medicine and Biology Society (EMBC)}, pp.\  3163--3166. IEEE,
  2016.

\bibitem[Bodla et~al.(2021)Bodla, Shrivastava, Chellappa, and
  Shrivastava]{bodla2021hierarchical}
Navaneeth Bodla, Gaurav Shrivastava, Rama Chellappa, and Abhinav Shrivastava.
\newblock Hierarchical video prediction using relational layouts for
  human-object interactions.
\newblock In \emph{Proceedings of the IEEE/CVF Conference on Computer Vision
  and Pattern Recognition}, pp.\  12146--12155, 2021.

\bibitem[Castellini et~al.(2020)Castellini, Bicego, Masillo, Zuccotto, and
  Farinelli]{Castellini2020TimeSS}
Alberto Castellini, Manuele Bicego, Francesco Masillo, Maddalena Zuccotto, and
  Alessandro Farinelli.
\newblock Time series segmentation for state-model generation of autonomous
  aquatic drones: A systematic framework.
\newblock \emph{Eng. Appl. Artif. Intell.}, 90:\penalty0 103499, 2020.

\bibitem[Deldari et~al.(2020)Deldari, Smith, Sadri, and
  Salim]{Deldari2020ESPRESSOEA}
Shohreh Deldari, Daniel~V. Smith, Amin Sadri, and Flora~D. Salim.
\newblock Espresso: Entropy and shape aware time-series segmentation for
  processing heterogeneous sensor data.
\newblock \emph{Proc. ACM Interact. Mob. Wearable Ubiquitous Technol.},
  4:\penalty0 77:1--77:24, 2020.

\bibitem[Denton \& Fergus(2018)Denton and Fergus]{denton2018stochastic}
Emily Denton and Rob Fergus.
\newblock Stochastic video generation with a learned prior.
\newblock In \emph{International Conference on Machine Learning}, pp.\
  1174--1183. PMLR, 2018.

\bibitem[Ermshaus et~al.(2022)Ermshaus, Sch{\"a}fer, and
  Leser]{ermshaus2022clasp}
Arik Ermshaus, Patrick Sch{\"a}fer, and Ulf Leser.
\newblock Clasp--parameter-free time series segmentation.
\newblock \emph{arXiv preprint arXiv:2207.13987}, 2022.

\bibitem[Fize et~al.(2017)Fize, Shrivastava, and M{\'e}nard]{fize2017geodict}
Jacques Fize, Gaurav Shrivastava, and Pierre~Andr{\'e} M{\'e}nard.
\newblock Geodict: an integrated gazetteer.
\newblock In \emph{Proceedings of Language, Ontology, Terminology and Knowledge
  Structures Workshop (LOTKS 2017)}, 2017.

\bibitem[Friedman et~al.(2008)Friedman, Hastie, and
  Tibshirani]{friedman2008sparse}
Jerome Friedman, Trevor Hastie, and Robert Tibshirani.
\newblock Sparse inverse covariance estimation with the graphical lasso.
\newblock \emph{Biostatistics}, 9\penalty0 (3):\penalty0 432--441, 2008.

\bibitem[Gharghabi et~al.(2019)Gharghabi, Yeh, Ding, Ding, Hibbing, LaMunion,
  Kaplan, Crouter, and Keogh]{gharghabi2019domain}
Shaghayegh Gharghabi, Chin-Chia~Michael Yeh, Yifei Ding, Wei Ding, Paul
  Hibbing, Samuel LaMunion, Andrew Kaplan, Scott~E Crouter, and Eamonn Keogh.
\newblock Domain agnostic online semantic segmentation for multi-dimensional
  time series.
\newblock \emph{Data mining and knowledge discovery}, 33\penalty0 (1):\penalty0
  96--130, 2019.

\bibitem[Hallac et~al.(2017)Hallac, Park, Boyd, and
  Leskovec]{hallac2017network}
David Hallac, Youngsuk Park, Stephen Boyd, and Jure Leskovec.
\newblock Network inference via the time-varying graphical lasso.
\newblock In \emph{Proceedings of the 23rd ACM SIGKDD International Conference
  on Knowledge Discovery and Data Mining}, pp.\  205--213, 2017.

\bibitem[Harguess \& Aggarwal(2009)Harguess and Aggarwal]{harguess2009semantic}
Josh Harguess and JK~Aggarwal.
\newblock Semantic labeling of track events using time series segmentation and
  shape analysis.
\newblock In \emph{2009 16th IEEE International Conference on Image Processing
  (ICIP)}, pp.\  4317--4320. IEEE, 2009.

\bibitem[Haury et~al.(2012)Haury, Mordelet, Vera-Licona, and
  Vert]{haury2012tigress}
Anne-Claire Haury, Fantine Mordelet, Paola Vera-Licona, and Jean-Philippe Vert.
\newblock {TIGRESS}: trustful inference of gene regulation using stability
  selection.
\newblock \emph{BMC systems biology}, 6\penalty0 (1), 2012.

\bibitem[Imani \& Keogh(2021)Imani and Keogh]{imani2021multi}
Shima Imani and E~Keogh.
\newblock Multi-window-finder: Domain agnostic window size for time series
  data, 2021.

\bibitem[Imani et~al.()Imani, Abdoli, and Keogh]{imanitime2cluster}
Shima Imani, Alireza Abdoli, and Eamonn Keogh.
\newblock Time2cluster: Clustering time series using neighbor information.

\bibitem[Jiao et~al.(2021)Jiao, Zhang, Liu, Yang, Hou, Li, and
  Tang]{jiao2021new}
Licheng Jiao, Ruohan Zhang, Fang Liu, Shuyuan Yang, Biao Hou, Lingling Li, and
  Xu~Tang.
\newblock New generation deep learning for video object detection: A survey.
\newblock \emph{IEEE Transactions on Neural Networks and Learning Systems},
  33\penalty0 (8):\penalty0 3195--3215, 2021.

\bibitem[Koller \& Friedman(2009)Koller and Friedman]{koller2009probabilistic}
Daphne Koller and Nir Friedman.
\newblock \emph{Probabilistic graphical models: principles and techniques}.
\newblock MIT press, 2009.

\bibitem[Kozey-Keadle et~al.(2011)Kozey-Keadle, Libertine, Lyden, Staudenmayer,
  and Freedson]{kozey2011validation}
Sarah Kozey-Keadle, Amanda Libertine, Kate Lyden, John Staudenmayer, and
  Patty~S Freedson.
\newblock Validation of wearable monitors for assessing sedentary behavior.
\newblock \emph{Medicine \& Science in Sports \& Exercise}, 43\penalty0
  (8):\penalty0 1561--1567, 2011.

\bibitem[Lan \& Sun(2015)Lan and Sun]{lan2015automated}
Rongyi Lan and Huaijiang Sun.
\newblock Automated human motion segmentation via motion regularities.
\newblock \emph{The Visual Computer}, 31:\penalty0 35--53, 2015.

\bibitem[Lin et~al.(2016)Lin, Karg, and Kuli{\'c}]{lin2016movement}
Jonathan Feng-Shun Lin, Michelle Karg, and Dana Kuli{\'c}.
\newblock Movement primitive segmentation for human motion modeling: A
  framework for analysis.
\newblock \emph{IEEE Transactions on Human-Machine Systems}, 46\penalty0
  (3):\penalty0 325--339, 2016.

\bibitem[Lu \& Huang(2020)Lu and Huang]{Lu2020SegmentationOM}
Shaowen Lu and Shuyu Huang.
\newblock Segmentation of multivariate industrial time series data based on
  dynamic latent variable predictability.
\newblock \emph{IEEE Access}, 8:\penalty0 112092--112103, 2020.

\bibitem[Machn{\'e} et~al.(2017)Machn{\'e}, Murray, and
  Stadler]{Machn2017SimilarityBasedSO}
Rainer Machn{\'e}, Douglas~B. Murray, and Peter~F. Stadler.
\newblock Similarity-based segmentation of multi-dimensional signals.
\newblock \emph{Scientific Reports}, 7, 2017.

\bibitem[Moerman et~al.(2019)Moerman, Aibar~Santos, Bravo Gonz{\'a}lez-Blas,
  Simm, Moreau, Aerts, and Aerts]{moerman2019grnboost2}
Thomas Moerman, Sara Aibar~Santos, Carmen Bravo Gonz{\'a}lez-Blas, Jaak Simm,
  Yves Moreau, Jan Aerts, and Stein Aerts.
\newblock Grnboost2 and arboreto: efficient and scalable inference of gene
  regulatory networks.
\newblock \emph{Bioinformatics}, 35\penalty0 (12):\penalty0 2159--2161, 2019.

\bibitem[Omranian et~al.(2015)Omranian, Mueller-Roeber, and
  Nikoloski]{omranian2015segmentation}
Nooshin Omranian, Bernd Mueller-Roeber, and Zoran Nikoloski.
\newblock Segmentation of biological multivariate time-series data.
\newblock \emph{Scientific reports}, 5\penalty0 (1):\penalty0 1--6, 2015.

\bibitem[Oprea et~al.(2020)Oprea, Martinez-Gonzalez, Garcia-Garcia,
  Castro-Vargas, Orts-Escolano, Garcia-Rodriguez, and Argyros]{oprea2020review}
Sergiu Oprea, Pablo Martinez-Gonzalez, Alberto Garcia-Garcia, John~Alejandro
  Castro-Vargas, Sergio Orts-Escolano, Jose Garcia-Rodriguez, and Antonis
  Argyros.
\newblock A review on deep learning techniques for video prediction.
\newblock \emph{IEEE Transactions on Pattern Analysis and Machine
  Intelligence}, 44\penalty0 (6):\penalty0 2806--2826, 2020.

\bibitem[Pu et~al.(2021)Pu, Cao, Zhang, Dong, and Chen]{pu2021learning}
Xingyue Pu, Tianyue Cao, Xiaoyun Zhang, Xiaowen Dong, and Siheng Chen.
\newblock Learning to learn graph topologies.
\newblock \emph{Advances in Neural Information Processing Systems}, 34, 2021.

\bibitem[Reinhardt et~al.(2013)Reinhardt, Christin, and
  Kanhere]{reinhardt2013predicting}
Andreas Reinhardt, Delphine Christin, and Salil~S Kanhere.
\newblock Predicting the power consumption of electric appliances through time
  series pattern matching.
\newblock In \emph{Proceedings of the 5th ACM Workshop on Embedded Systems For
  Energy-Efficient Buildings}, pp.\  1--2, 2013.

\bibitem[Reiss \& Stricker(2012)Reiss and Stricker]{reiss2012introducing}
Attila Reiss and Didier Stricker.
\newblock Introducing a new benchmarked dataset for activity monitoring.
\newblock In \emph{2012 16th international symposium on wearable computers},
  pp.\  108--109. IEEE, 2012.

\bibitem[Roche et~al.(2017)Roche, Teisseire, and
  Shrivastava]{roche2017valorcarn}
Mathieu Roche, Maguelonne Teisseire, and Gaurav Shrivastava.
\newblock Valorcarn-tetis: Terms extracted with biotex.
\newblock 2017.

\bibitem[Rolfs et~al.(2012)Rolfs, Rajaratnam, Guillot, Wong, and
  Maleki]{rolfs2012iterative}
Benjamin Rolfs, Bala Rajaratnam, Dominique Guillot, Ian Wong, and Arian Maleki.
\newblock Iterative thresholding algorithm for sparse inverse covariance
  estimation.
\newblock \emph{Advances in Neural Information Processing Systems},
  25:\penalty0 1574--1582, 2012.

\bibitem[Saini et~al.(2022)Saini, He, Shrivastava, Rambhatla, and
  Shrivastava]{saini2022recognizing}
Nirat Saini, Bo~He, Gaurav Shrivastava, Sai~Saketh Rambhatla, and Abhinav
  Shrivastava.
\newblock Recognizing actions using object states.
\newblock In \emph{ICLR2022 Workshop on the Elements of Reasoning: Objects,
  Structure and Causality}, 2022.

\bibitem[Sch{\"a}fer et~al.(2021)Sch{\"a}fer, Ermshaus, and
  Leser]{Schfer2021ClaSPT}
Patrick Sch{\"a}fer, Arik Ermshaus, and Ulf Leser.
\newblock Clasp - time series segmentation.
\newblock \emph{Proceedings of the 30th ACM International Conference on
  Information \& Knowledge Management}, 2021.

\bibitem[Serra et~al.(2014)Serra, M{\"u}ller, Grosche, and
  Arcos]{serra2014unsupervised}
Joan Serra, Meinard M{\"u}ller, Peter Grosche, and Josep~Ll Arcos.
\newblock Unsupervised music structure annotation by time series structure
  features and segment similarity.
\newblock \emph{IEEE Transactions on Multimedia}, 16\penalty0 (5):\penalty0
  1229--1240, 2014.

\bibitem[Shrivastava(2021)]{shrivastava2021diversethesis}
Gaurav Shrivastava.
\newblock Diverse video generation.
\newblock Master's thesis, University of Maryland, College Park, 2021.

\bibitem[Shrivastava \& Shrivastava(2021)Shrivastava and
  Shrivastava]{shrivastava2021diverse}
Gaurav Shrivastava and Abhinav Shrivastava.
\newblock Diverse video generation using a gaussian process trigger.
\newblock \emph{arXiv preprint arXiv:2107.04619}, 2021.

\bibitem[Shrivastava \& Shrivastava(2024)Shrivastava and
  Shrivastava]{shrivastava2024video1}
Gaurav Shrivastava and Abhinav Shrivastava.
\newblock Video prediction by modeling videos as continuous multi-dimensional
  processes.
\newblock In \emph{Proceedings of the IEEE/CVF Conference on Computer Vision
  and Pattern Recognition}, pp.\  7236--7245, 2024.

\bibitem[Shrivastava et~al.(2024{\natexlab{a}})Shrivastava, Lim, and
  Shrivastava]{shrivastava2024video2}
Gaurav Shrivastava, Ser-Nam Lim, and Abhinav Shrivastava.
\newblock Video decomposition prior: Editing videos layer by layer.
\newblock In \emph{The Twelfth International Conference on Learning
  Representations}, 2024{\natexlab{a}}.

\bibitem[Shrivastava et~al.(2024{\natexlab{b}})Shrivastava, Lim, and
  Shrivastava]{shrivastava2024video3}
Gaurav Shrivastava, Ser~Nam Lim, and Abhinav Shrivastava.
\newblock Video dynamics prior: An internal learning approach for robust video
  enhancements.
\newblock \emph{Advances in Neural Information Processing Systems}, 36,
  2024{\natexlab{b}}.

\bibitem[Shrivastava(2020)]{shrivastava2020using}
Harsh Shrivastava.
\newblock \emph{On Using Inductive Biases for Designing Deep Learning
  Architectures}.
\newblock PhD thesis, Georgia Institute of Technology, 2020.

\bibitem[Shrivastava \& Chajewska(2022{\natexlab{a}})Shrivastava and
  Chajewska]{shrivastava2022methods}
Harsh Shrivastava and Urszula Chajewska.
\newblock Methods for recovering conditional independence graphs: A survey.
\newblock \emph{arXiv preprint arXiv:2211.06829}, 2022{\natexlab{a}}.

\bibitem[Shrivastava \& Chajewska(2022{\natexlab{b}})Shrivastava and
  Chajewska]{shrivastava2022neural}
Harsh Shrivastava and Urszula Chajewska.
\newblock Neural graphical models.
\newblock \emph{arXiv preprint arXiv:2210.00453}, 2022{\natexlab{b}}.

\bibitem[Shrivastava \& Chajewska(2023)Shrivastava and
  Chajewska]{shrivastava2023neural}
Harsh Shrivastava and Urszula Chajewska.
\newblock Neural graph revealers.
\newblock \emph{arXiv preprint arXiv:2302.13582}, 2023.

\bibitem[Shrivastava \& Chajewska(2024)Shrivastava and
  Chajewska]{shrivastava2024methods}
Harsh Shrivastava and Urszula Chajewska.
\newblock Methods for recovering conditional independence graphs: a survey.
\newblock \emph{Journal of Artificial Intelligence Research}, 80:\penalty0
  593--612, 2024.

\bibitem[Shrivastava et~al.(2019)Shrivastava, Chen, Chen, Lan, Aluru, Liu, and
  Song]{shrivastava2019glad}
Harsh Shrivastava, Xinshi Chen, Binghong Chen, Guanghui Lan, Srinvas Aluru, Han
  Liu, and Le~Song.
\newblock \texttt{GLAD}: Learning sparse graph recovery.
\newblock \emph{arXiv preprint arXiv:1906.00271}, 2019.

\bibitem[Shrivastava et~al.(2020)Shrivastava, Zhang, Aluru, and
  Song]{shrivastava2020grnular}
Harsh Shrivastava, Xiuwei Zhang, Srinivas Aluru, and Le~Song.
\newblock Grnular: Gene regulatory network reconstruction using unrolled
  algorithm from single cell rna-sequencing data.
\newblock \emph{bioRxiv}, 2020.

\bibitem[Shrivastava et~al.(2022{\natexlab{a}})Shrivastava, Chajewska, Abraham,
  and Chen]{shrivastava2022a}
Harsh Shrivastava, Urszula Chajewska, Robin Abraham, and Xinshi Chen.
\newblock A deep learning approach to recover conditional independence graphs.
\newblock In \emph{NeurIPS 2022 Workshop: New Frontiers in Graph Learning},
  2022{\natexlab{a}}.
\newblock URL \url{https://openreview.net/forum?id=kEwzoI3Am4c}.

\bibitem[Shrivastava et~al.(2022{\natexlab{b}})Shrivastava, Chajewska, Abraham,
  and Chen]{shrivastava2022uglad}
Harsh Shrivastava, Urszula Chajewska, Robin Abraham, and Xinshi Chen.
\newblock \texttt{uGLAD}: Sparse graph recovery by optimizing deep unrolled
  networks.
\newblock \emph{arXiv preprint arXiv:2205.11610}, 2022{\natexlab{b}}.

\bibitem[Shrivastava et~al.(2022{\natexlab{c}})Shrivastava, Zhang, Song, and
  Aluru]{shrivastava2022grnular}
Harsh Shrivastava, Xiuwei Zhang, Le~Song, and Srinivas Aluru.
\newblock Grnular: A deep learning framework for recovering single-cell gene
  regulatory networks.
\newblock \emph{Journal of Computational Biology}, 29\penalty0 (1):\penalty0
  27--44, 2022{\natexlab{c}}.

\bibitem[V{\^a}n Anh Huynh-Thu et~al.(2010)V{\^a}n Anh Huynh-Thu, Wehenkel, and
  Geurts]{van2010inferring}
Alexandre~Irrthum V{\^a}n Anh Huynh-Thu, Louis Wehenkel, and Pierre Geurts.
\newblock Inferring regulatory networks from expression data using tree-based
  methods.
\newblock \emph{PloS one}, 5\penalty0 (9), 2010.

\bibitem[Yeh et~al.(2016)Yeh, Zhu, Ulanova, Begum, Ding, Dau, Silva, Mueen, and
  Keogh]{yeh2016matrix}
Chin-Chia~Michael Yeh, Yan Zhu, Liudmila Ulanova, Nurjahan Begum, Yifei Ding,
  Hoang~Anh Dau, Diego~Furtado Silva, Abdullah Mueen, and Eamonn Keogh.
\newblock Matrix profile i: all pairs similarity joins for time series: a
  unifying view that includes motifs, discords and shapelets.
\newblock In \emph{2016 IEEE 16th international conference on data mining
  (ICDM)}, pp.\  1317--1322. Ieee, 2016.

\bibitem[Zhou et~al.(2022)Zhou, Porikli, Crandall, Van~Gool, and
  Wang]{zhou2022survey}
Tianfei Zhou, Fatih Porikli, David~J Crandall, Luc Van~Gool, and Wenguan Wang.
\newblock A survey on deep learning technique for video segmentation.
\newblock \emph{IEEE transactions on pattern analysis and machine
  intelligence}, 45\penalty0 (6):\penalty0 7099--7122, 2022.

\end{thebibliography}
\bibliographystyle{iclr2023_conference}

\clearpage
% \appendix

\end{document}